\documentclass[sigconf]{acmart}
\usepackage{graphicx}
\usepackage{multirow}
\usepackage{algorithm}
\usepackage[capitalize]{cleveref}
\usepackage{subcaption}
\usepackage{amsmath}    
\usepackage{enumitem}

\AtBeginDocument{%
  }

\copyrightyear{2025}
\acmYear{2025}
\setcopyright{acmlicensed}\acmConference[MMAsia '25]{ACM Multimedia Asia}{December 9--12, 2025}{Kuala Lumpur, Malaysia}
\acmBooktitle{ACM Multimedia Asia (MMAsia '25), December 9--12, 2025, Kuala Lumpur, Malaysia}
\acmDOI{10.1145/3743093.3771031}
\acmISBN{979-8-4007-2005-5/2025/12}

\begin{document}

\title{SELECT: Detecting Label Errors in Real-world Scene Text Data}

\author{Wenjun Liu}
% \email{liuwenjun6146@gmail.com}
\affiliation{%
  \institution{Yidun AI Lab, NetEase}
  \city{Hangzhou}
  \country{China}
}

\author{Qian Wu}
% \email{wuqianwq11@gmail.com}
\affiliation{%
  \institution{Yidun AI Lab, NetEase}
  \city{Hangzhou}
  \country{China}
}

\author{Yifeng Hu}
% \email{huyifeng1188@gmail.com}
\affiliation{%
  \institution{Yidun AI Lab, NetEase}
  \city{Hangzhou}
  \country{China}
}

\author{Yuke Li}
% \email{liyuke@corp.netease.com}
\affiliation{%
  \institution{Yidun AI Lab, NetEase}
  \city{Hangzhou}
  \country{China}
}
\authornote{Corresponding author.}

% \author{Wenjun Liu, Qian Wu, Yifeng Hu, Yuke Li}
% % \authornote{Both authors contributed equally to this research.}
% % \email{liyuke@corp.netease.com}
% \affiliation{%
%   \institution{Yidun AI Lab, NetEase}
%   \city{Hangzhou}
%   \country{China}
% }

\renewcommand{\shortauthors}{Wenjun Liu et al.}

\begin{abstract}
  % We introduce SELECT (\textbf{S}cene t\textbf{E}xt \textbf{L}abel \textbf{E}rrors dete\textbf{CT}ion), a novel approach that leverages multi-modal training to detect label errors in real-world scene text datasets. Utilizing an image-text encoder and a character-level tokenizer, SELECT addresses variable-length sequence labels, label sequence misalignment, and character-level errors, outperforming existing methods in accuracy and practical utility. In addition, we introduce Similarity-based Sequence Label Corruption (SSLC), a process that intentionally introduces errors into the training labels to mimic real-world error scenarios during training. SSLC not only can cause a change in the sequence length but also takes into account the visual similarity between characters during corruption. Our method is the first to detect label errors in real-world scene text datasets successfully. Experimental results demonstrate the effectiveness of SELECT in detecting label errors and improving STR accuracy on real-world text datasets, showcasing its practical utility.

  We introduce SELECT (\textbf{S}cene t\textbf{E}xt \textbf{L}abel \textbf{E}rrors dete\textbf{CT}ion), a novel approach that leverages multi-modal training to detect label errors in real-world scene text datasets. Utilizing an image-text encoder and a character-level tokenizer, SELECT addresses the issues of variable-length sequence labels, label sequence misalignment, and character-level errors, outperforming existing methods in accuracy and practical utility. In addition, we introduce Similarity-based Sequence Label Corruption (SSLC), a process that intentionally introduces errors into the training labels to mimic real-world error scenarios during training. SSLC not only can cause a change in the sequence length but also takes into account the visual similarity between characters during corruption. Our method is the first to detect label errors in real-world scene text datasets successfully accounting for variable-length labels. Experimental results demonstrate the effectiveness of SELECT in detecting label errors and improving STR accuracy on real-world text datasets, showcasing its practical utility.

\end{abstract}





\maketitle

\section{Introduction}
\label{sec:intro}

\textbf{Scene Text Recognition (STR)} is crucial for applications like autonomous vehicles and intelligent surveillance. Research shows that STR models trained on real data outperform those trained on synthetic data\cite{baek2021what}. However, real data often contains annotation noise, harming STR accuracy. 
Our analysis indicates that removing about 8.7\% potential noisy data from the real dataset would enhance the STR performance, highlighting the necessity of \textbf{label error detection (LED)}.
% Our analysis reveals that about 8.7\% of real dataset labels are noisy, stressing the need for effective \textbf{label error detection (LED)}.

% Despite recent progress in LED methods \cite{northcutt2021confidentlearning, schubert2023objectdetection1} for various tasks, detecting label errors in STR remains challenging due to variable-length sequence labels, character-level errors and label sequence misalignment when insertion or deletion occurs (as in \cref{fig:SELECT_results} (a,b)). This misalignment, renders existing approaches based on strict token-to-token \cite{northcutt2021confidentlearning} matching between predicted and annotated labels unsuitable for scene text data. To date, the existing literature on STR noise detection \cite{liu2021confidentsquencelearning} ignores the label sequence misalignment issues and simply truncates the tokens in the prediction to match the length of the sequence label and corresponds them one by one with the labels. This truncation approach fails to account for the nuances of spatial alignment and contextual meaning between the predicted text and the actual labels, which is crucial for accurately identifying insertion and deletion errors in this task. Consequently, its application is confined to synthetic datasets and cannot effectively handle noise issues in real-world data.

\begin{figure}[t]
    \centering
    \includegraphics[width=\linewidth]{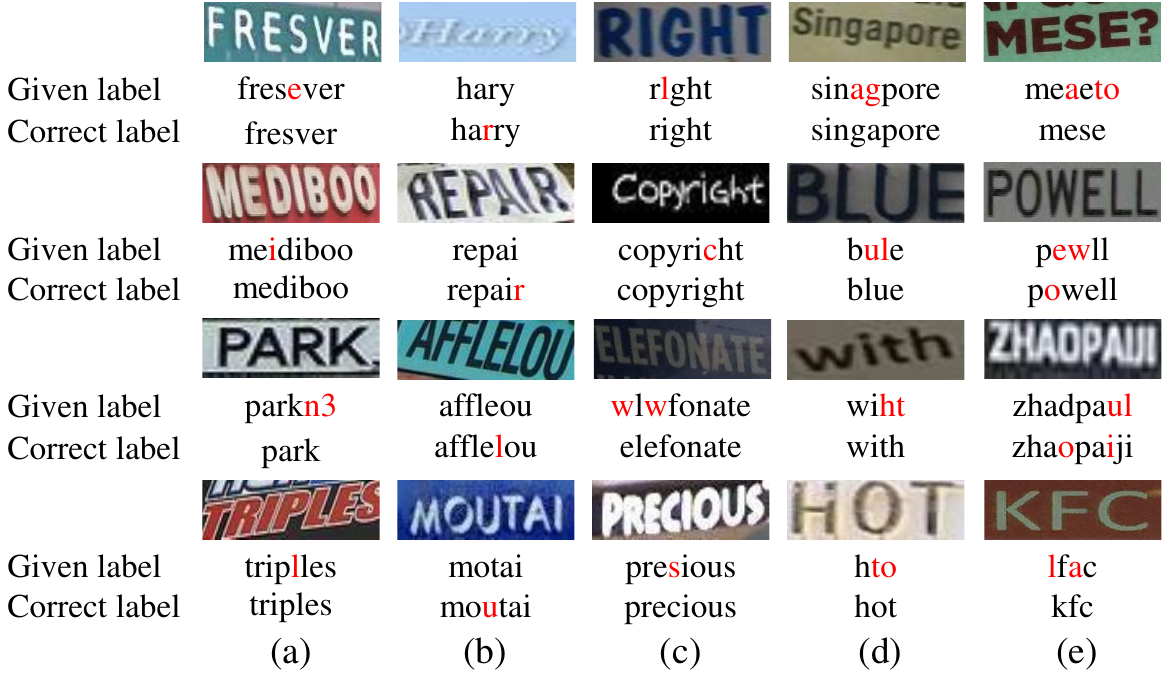}
    % \vspace{-0.8em}
    % \caption{Visualization of samples with label errors detected by SELECT. From left to right: (a) insertion, (b) deletion, (c) substitution, (d) transposition, and (e) multiple types of noise. The incorrect or deleted characters in the labels are highlighted in red.}
    \caption{Some samples with error labels in real datasets detected by SELECT. Sequence noise can be classified as: (a) insertion, (b) deletion, (c) substitution, (d) transposition, and (e) multiple types of noise. The incorrect or deleted characters in the labels are highlighted in red.}
    \Description{Some samples with error labels in real datasets detected by SELECT. Sequence noise can be classified as: (a) insertion, (b) deletion, (c) substitution, (d) transposition, and (e) multiple types of noise. The incorrect or deleted characters in the labels are highlighted in red.}    
    \label{fig:SELECT_results}
\end{figure}

Despite recent progress in LED methods \cite{northcutt2021confidentlearning, schubert2023objectdetection1} for various tasks, detecting label errors in STR remains challenging due to variable-length sequence labels, character-level errors, and label sequence misalignment when insertion or deletion occurs (as shown in \cref{fig:SELECT_results} (a,b)). This misalignment, renders existing approaches based on strict token-to-token \cite{northcutt2021confidentlearning} matching between predicted and annotated labels unsuitable for scene text data. To date, the existing literature on STR noise detection \cite{liu2021confidentsquencelearning} ignores the label sequence misalignment issues and truncates the tokens in the prediction to match the length of the sequence label and correspond them one by one with the labels. This truncation approach fails to account for the nuances of spatial alignment and contextual meaning between the predicted text and the actual labels, which is crucial for accurately identifying insertion and deletion errors in this task. Consequently, its application is confined to synthetic datasets and cannot effectively handle noise issues in real-world data.

Building upon the limitations of existing approaches, we propose a novel method called \textbf{SELECT} (\textbf{S}cene t\textbf{E}xt \textbf{L}abel \textbf{E}rrors dete\textbf{CT}ion).  Our approach involves performing vision-language training on synthetic datasets to detect label errors on real datasets. By directly inputting an image and its corresponding label sequence into an image-text encoder, we bypass the need to align predicted and annotated labels, thus accommodating variable-length text labels and resolving label sequence misalignment issues. Each text character acts as a separate query in the image-text encoder, capturing precise character information and location from the image features. To train the SELECT model, we incorporate the \textbf{Similarity-based Sequence Label Corruption (SSLC)}, which generates negative samples by corrupting sequence labels through operations such as insertion, substitution, deletion, and transposition during training. In addition, we leverage the visual similarity between characters to capture the confusion that annotators may experience when labeling visually similar characters. The main contributions of our work are as follows:
\begin{itemize}
    % \item We identify that 8.7\% of real-world scene text data contains label noise, stressing the need for LED in STR.

    \item We propose a novel method, SELECT, which leverages image-text training on synthetic datasets to detect label errors in real-world scene text data. To our knowledge, we are the first to explore LED in real-world scene text accounting for variable-length labels.
    
    \item We design the SSLC, which equips SELECT to handle sequence label errors by considering label length and visual character similarity.
    
    \item Our experiments validate the effectiveness of SELECT in LED with a high accuracy of 98.45\% in synthetic data. Furthermore, we demonstrate significant improvements in STR models trained on clean real data with noisy data removed by SELECT, even with an 8.7\% reduction. 
    
\end{itemize}

% \begin{figure}[t]
%     \centering
%     \includegraphics[width=0.5\linewidth]{noise_examples.pdf}
%     \caption{Some samples with error labels in real datasets. Sequence noise can be classified into four categories \cite{brill2000improved}: (a) insertion, (b) deletion, (c) substitution, and (d) transposition. The incorrect or deleted characters in the labels are highlighted in red.}
%     \label{fig:noise_examples}
%     \vspace{-8pt}
% \end{figure}

\section{LITERATURE REVIEW}
\label{sec:related}

\smallskip
\textbf{Scene Text Recognition.}
Numerous methods have been proposed for STR, including Connectionist Temporal Classification, which maps images to character sequences, and attention mechanisms \cite{zhang2023linguisticmore} aligning input information with output features. Recently, transformers have shown promise. For instance, ViTSTR \cite{atienza2021vitstr} uses the Vision Transformer (ViT) \cite{dosovitskiy2020vit}, and PARSeq \cite{bautista2022parseq} employs PLM \cite{yang2019xlnet} to reorder characters for predictions. However, label errors in real-world scene text pose a challenge. While some papers \cite{zhang2023linguisticmore} have addressed attention drift, they overlook the impact of annotation errors. 

\textbf{Label Error Detection (LED)}
is gaining attention across various tasks, such as classification \cite{northcutt2021confidentlearning}, object detection \cite{schubert2023objectdetection1}, and semantic segmentation. However, these methods often align predicted and annotated token counts, which is unsuitable for STR due to variable-length labels and sequence misalignment. Only one relevant work \cite{liu2021confidentsquencelearning} applies Confident Learning \cite{northcutt2021confidentlearning} to detect erroneous tokens in STR. This study, using manually corrupted synthetic datasets, addresses substitution and transposition errors, ignoring insertion and deletion errors that alter sequence length, thus limiting its effectiveness for real data. There is a clear need for novel, tailored approaches for effective LED in real-world STR datasets.

\begin{figure*}[!tp]
    \centering
    \includegraphics[width=0.98\linewidth]{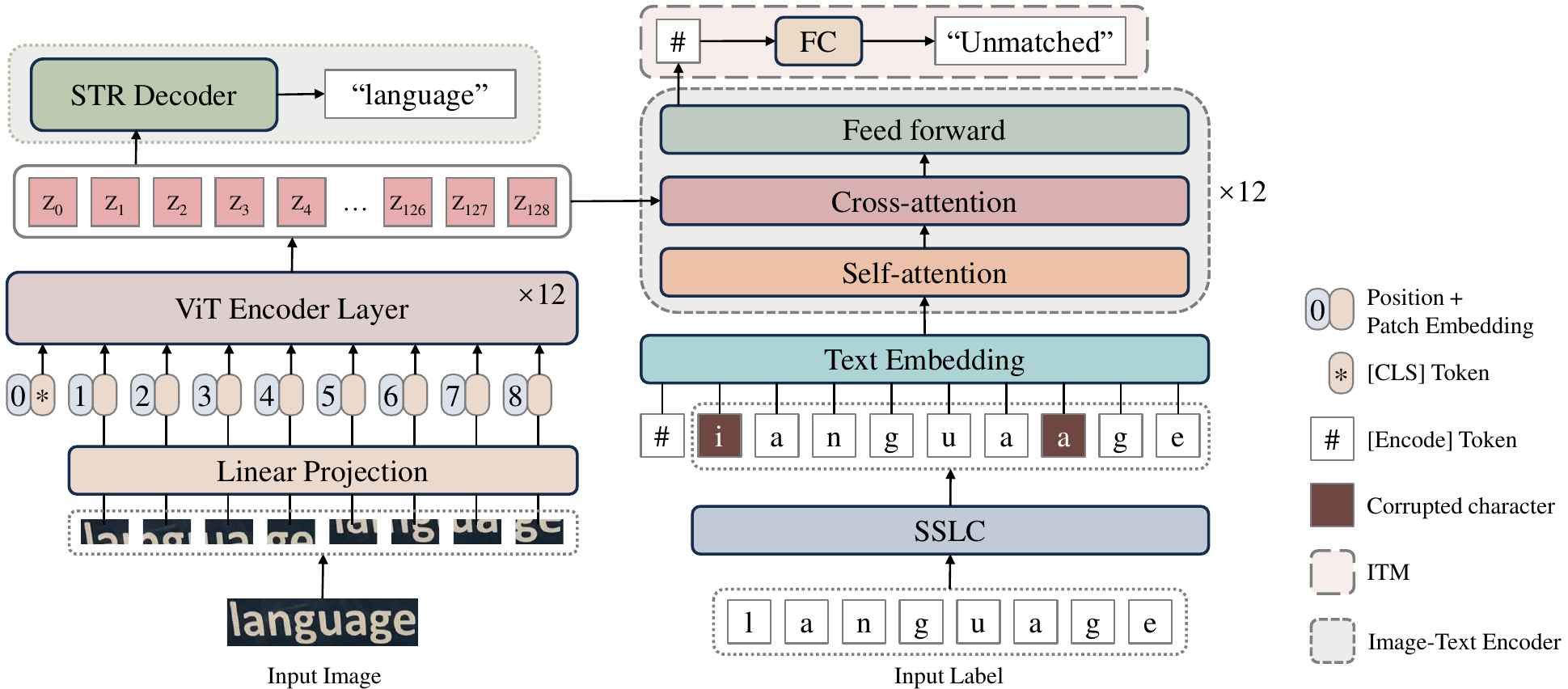}
    \caption{The proposed SELECT. During training, SSLC (Similarity-based Sequence Label Corruption) corrupts the original label to obtain a corrupted label, using both the corrupted label and the original image as a negative sample pair for training SELECT. During inference, the original label and image are input into SELECT to determine whether they match. The auxiliary learning of the STR decoder is used on the head of the ViT encoder.}
    \Description{The proposed SELECT. During training, SSLC (Similarity-based Sequence Label Corruption) corrupts the original label to obtain a corrupted label, using both the corrupted label and the original image as a negative sample pair for training SELECT. During inference, the original label and image are input into SELECT to determine whether they match. The auxiliary learning of the STR decoder is used on the head of the ViT encoder.}
    % \vspace{-1.5em}
    \label{fig:select}
\end{figure*}

\smallskip
\textbf{Vision-language Pre-training (VLP)}
trains models for integrated vision and language tasks. To effectively train VLP models, researchers % \cite{cho2021vlp3, chen2022vlp4, li2022blip, wang2022vlp5} 
commonly use objectives such as Image-Text Contrastive learning (ITC), Image-Text Matching (ITM), and Language Modeling (LM). Recently, VLP models have been applied to STR. OFASTR \cite{lin2022ofastr} treated STR as an image captioning task, fine-tuning the VLP model, to improve the recognition results. CLIP4STR \cite{zhao2023clip4str} utilized the image and text encoders of CLIP \cite{radford2021vlp11} to extract modal features, fused for text prediction. Notably, no existing work has applied VLP for scene text LED. VLP presents an opportunity to enhance LED by leveraging rich contextual information encoded in pre-trained models, thereby improving scene text comprehension.

\section{Method}
\label{sec:method}

% In this section, we first present the architecture of SELECT (\textbf{S}cene t\textbf{E}xt \textbf{L}abel \textbf{E}rrors dete\textbf{CT}ion). Next, we demonstrate the SSLC (Similarity-based Sequence Label Corruption) process of generating synthetic label errors to simulate real-world conditions.

\subsection{Scene Text Label Error Detection}

SELECT, employs a dual-encoder setup as shown in \cref{fig:select}, enhancing LED by leveraging both visual and textual information. The multi-modal image-text encoder with Multi-Head Cross-Attention module (MHCA) emphasizes areas of the image corresponding to individual characters, with the final output $p^{itm}$ representing the likelihood of a match between the image and its text label.

\smallskip
\textbf{Visual Encoder.} ViT \cite{dosovitskiy2020vit} is utilized as the image encoder $Enc_I$ to segment the scene text image into patches and encoding them into embeddings $Z = \{z_i\}^M_{i=0}$, where $z_i (i > 0)$ is the $i$-th patch embeddings, $M$ is the total patch number and $z_0$ is the specific [CLS] token to capture global features. % To handle scene text images with a horizontal rectangular shape, we adapt the input image size to $64\times256$ and modify the patch size to $8\times16$.

\smallskip
\textbf{Image-Text Encoder.}
Our image-text encoder $Enc_{IT}$ is inspired by the image-grounded text encoder in BLIP \cite{li2022blip} but with a character-level tokenizer. %, which enhances sensitivity to text data anomalies and plays a key role in label errors detection. 
The inputs of $Enc_{IT}$ are the image embeddings $Z$ 
%encoded by the visual encoder $Enc_I$ 
and the tokenized text label embeddings $T=\{t_i\}^n_{i=0}$ of the image, where $t_i (i > 0)$ is the embedding of the i-th character in text label $L$, $t_0$ is an [Encode] token and $n$ is the actual length of a given sequence label $L$. 
%Note that the text label $L$ can be variable-length, e.g., $n$ can be any integer from $1$ to max label length $N$. 
% Similar to the [CLS] token used in $Enc_I$, 
The [Encode] token is designed to aggregate global contextual features from both text and image and aid in identifying label discrepancies. 

% GPT from wq
The $Enc_{IT}$ comprises 12 transformer blocks, integrating and refining visual and textual embeddings. Initially, the MHSA operation enhances text embeddings by focusing on token orders and relations:
\begin{align}
T' = MHSA(T) + T
\end{align}
%where $T$ represents input text embeddings.
Subsequently, the MHCA mechanism  allows these text embeddings $T'$ to interact with visual embeddings $Z$:
\begin{align}
T'' = MHCA(T', Z) + T'
\end{align}
Finally, the Feed Forward Network (FFN) is applied to each embedding individually:
\begin{align}
T''' = FFN(T''),
\end{align}
yielding the final output $T'''$ of the block, serving as the input to the next block or as the final representation. % if it's the last block. 
The embedding generated by the [Encode] token, serving as the comprehensive image-text representation $f_{IT}$ and providing a holistic view of the image-label pair, is crucial for the final error detection.
% The [Encode] token's generated embedding $f_{IT}$ provides a comprehensive image-text representation crucial for the final error detection. % phase.

% \paragraph{ITM} To effectively align images with text labels and discern their match quality, we employ the Image-Text Matching loss (ITM) as our binary classification objective. The output embedding $f_{IT}$, derived from the [Encode] token, encapsulates a nuanced, multi-modal representation that integrates visual and textual cues, thereby providing a robust basis for evaluating the congruence of image-label pairs. Next, a fully-connected layer and softmax function are applied to generate a two-class probability prediction $p^{itm}$. The ITM loss is as following:
% \begin{align}
% \mathcal{L}_{\text{ITM}} = \sum_{i=1}^2 y_i^{itm} \log p_i^{itm}
% \end{align}
% where $y^{itm}$ is the one-hot ground truth of whether an image-label pair is matched, and $i$ refers to the class index. The ITM loss $\mathcal{L}_{ITM}$ refines model predictions by emphasizing accurate classification of matched versus unmatched pairs, promoting fine-grained learning distinctions. By assessing the congruence of image-label pairs through the ITM framework, our model effectively navigates challenges such as label sequence misalignment and the presence of various noise types in the data. This capacity ensures robust error detection across diverse and unpredictable real-world datasets.

% GPT

\smallskip
\textbf{ITM.} To effectively align images with text labels and discern their match quality, we use the Image-Text Matching loss (ITM) for binary classification. %The embedding $f_{IT}$ from the [Encode] token integrates visual and textual cues, forming a basis for evaluating image-label congruence. A fully-connected layer and softmax function produce the two-class probability prediction $p^{itm}$. 
A fully connected layer and softmax function produce the two-class probability prediction $p^{itm}$ with the embedding $f_{IT}$ as input, which integrates visual and textual cues.
The ITM loss is represented as:
\begin{align}
\mathcal{L}_{\text{ITM}} = \sum_{i=1}^2 y_i^{itm} \log p_i^{itm}
\end{align}
where, $y^{itm}$ denotes whether an image-label pair is matched, and $i$ is the class index. The ITM loss $\mathcal{L}_{ITM}$ enhances model predictions by focusing on the accurate classification of matched versus unmatched pairs, facilitating detailed learning distinctions. Through the ITM framework, our model effectively addresses challenges like label sequence misalignment and diverse noise types in the data, ensuring robust error detection across unpredictable real-world datasets.

% original
% \paragraph{Tokenizer} BLIP uses WordPiece \cite{devlin2018bert} tokenizer, which breaks down words into subword units. This tokenizer has gained popularity in natural language processing tasks and has been employed in multiple STR studies \cite{li2023trocr,lin2022ofastr,fujitake2023dtrocr,wang2022multigranularity} to fully exploit the capabilities of pre-trained language models. However, it is not well-suited for this task as the tokenizer often groups multiple characters together. This dilutes focus on individual characters—especially in noisy conditions—and complicates learning by generating vastly different tokenizations for slight variations in input (e.g., noise in a single character). Therefore, we replace the original WordPiece tokenizer with a character-level tokenizer. This tokenizer treats each character as a separate token, allowing each character in the label to be located and extract relevant information from the image features, enabling the model to determine whether the character is correct.

\smallskip
\textbf{Tokenizer.} 
The WordPiece \cite{devlin2018bert} tokenizer used in BLIP, breaks down words into subword units, has gained popularity in natural language processing tasks, and has been employed in multiple STR studies \cite{lin2022ofastr}.
However, it is not well-suited for this task as the tokenizer often groups multiple characters. This dilutes focus on individual characters—especially in noisy conditions—and complicates learning by generating vastly different tokenizations for slight variations in input (e.g., noise in a single character). Therefore, we replace the original WordPiece tokenizer with a character-level tokenizer. This tokenizer treats each character as a separate token, allowing each character in the label to be located and extract relevant information from the image features, enabling the model to determine whether the character is correct.

\smallskip
\textbf{Auxiliary Learning.} In our SELECT method, the auxiliary learning component features a STR decoder head, utilizing a three-layer transformer decoder block with cross-attention mechanisms. A linear layer followed by softmax is used for final text prediction $p^{str}$. We use the auto-regressive decoding strategy to predict each character sequentially similar to \cite{lu2021master}, which harnesses contextual dependencies between characters in a label. The STR loss $\mathcal{L}_{STR}$ is shown as following:
\begin{align}
\mathcal{L}_{STR} = -\frac{1}{N} \sum_{i=1}^{N} \sum_{c=1}^{C} y_{i,c}^{str} \log(p_{i,c}^{str})
\end{align}
where $C$ is the character set and $y_{i,c}^{str}$ is the ground truth of i-th character. Integrating the STR decoder head significantly enhances text detection and positioning in images, thereby improving overall label error detection through a more detailed textual context analysis. Thus, the total loss is:
\begin{align}
\mathcal{L} = \mathcal{L}_{ITM} + \alpha\mathcal{L}_{STR} 
\end{align}
where $\alpha$ is the trade-off parameter, set as $1.0$ for default.

\subsection{Similarity-based Sequence Label Corruption}

We use image-label pairs from synthetic datasets to train the SELECT model. Positive samples are directly taken from the synthetic datasets, while negative samples are generated by corrupting the labels of positive pairs. Commonly employed methods for label corruption include randomly flipping the labels of clean data \cite{han2018coteaching}, either uniformly (symmetric) or non-uniformly (asymmetric), which involves randomly or artificially setting the noise transition matrix. However, these methods fall short in simulating the noise found in real data and are only able to model substitution noise types when it comes to sequence labels.
To address the issues above, we propose Similarity-based Sequence Label Corruption (SSLC) for resembling real-world noise in the STR dataset. SSLC employs a set of label corruption operations, including insertion, deletion, substitution, and transposition, to introduce noises into sequence labels. To simulate the likelihood of annotators confusing visually similar characters during annotation, SSLC builds a noise transition matrix derived from visual similarity as the foundation for substitution operation.

\smallskip
\textbf{Label Corruption Operations.} Given a character set (charset) $C$, a sequence label \(L\) can be represented as follows:
\begin{align}
L = [l_1, \ldots, l_{i-1},l_i,l_{i+1}, \ldots l_n]
\end{align}
where $l_i \in C$, $n \le N$, $N \in \mathbb{Z}$ is the max length of sequence labels and $n$ is the actual length of a given sequence label $L$. After undergoing one of the corruption operations individually, the resulting corrupted sequence label $\Tilde{L}$ can be expressed as:

% \begin{itemize}
%     \item Insertion: $\Tilde{L} = [l_1, \ldots, l_{i-1},l_i,\Tilde{l}_{j},l_{i+1}, \ldots, l_n] $,
%     where \(j \in \mathbb{Z}\) and \(1 \leq j \leq n+1\). 
%     % After insertion, the length of the sequence label increases by one, from $n$ to $n+1$.
%     \item Deletion: $ \Tilde{L} = [l_1, \ldots, l_{i-1},l_{i+1}, \ldots, l_n] $. 
%     % After deletion, the length of the sequence label decreases from $n$ to $n-1$.
%     \item Substitution: $ \Tilde{L} = [l_1, \ldots, l_{i-1},\Tilde{l}_i,l_{i+1}, \ldots l_n] $, where \(\Tilde{l}_i \in C \setminus \{l_i\} \).
%     \item Transposition: $\Tilde{L} = [l_1, \ldots, l_i, l_{i-1},l_{i+1}, \ldots l_n] $.
% \end{itemize}

\begin{itemize}
    \item Insertion: $\Tilde{L} = [l_1, \ldots, l_{i-1},l_i,\Tilde{l}_{j},l_{i+1}, \ldots, l_n] $,
    where \(j \in \mathbb{Z}\) and \(1 \leq j \leq n+1\). After insertion, the length of the sequence label increases by one, from $n$ to $n+1$.
    \item Deletion: $ \Tilde{L} = [l_1, \ldots, l_{i-1},l_{i+1}, \ldots, l_n] $. After deletion, the length of the sequence label decreases from $n$ to $n-1$.
    \item Substitution: $ \Tilde{L} = [l_1, \ldots, l_{i-1},\Tilde{l}_i,l_{i+1}, \ldots l_n] $, where \(\Tilde{l}_i \in C \setminus \{l_i\} \).
    \item Transposition: $\Tilde{L} = [l_1, \ldots, l_i, l_{i-1},l_{i+1}, \ldots l_n] $.
\end{itemize}
where insertion and deletion can alter the length of the label.

\smallskip
\textbf{Similarity-based Noise Transition Matrix.} A pivotal aspect of SSLC is capturing the confusion that annotators experience when labeling visually similar characters. We leverage the concept of visual similarity between characters to prioritize certain characters over others in the substitution process. For example, when replacing ``i''  during the substitution process, we prioritize ``l'' over ``n'' due to their visual similarity. To achieve this prioritization, we construct a noise transition matrix that quantifies the visual similarity between all characters in the charset based on their image features from synthetic images.
% original
% Firstly, we synthesize multiple images of a single character using various fonts and angles, and, in the case of English letters, both uppercase and lowercase formats are included. Secondly, a pre-trained ResNet50 \cite{he2016resnet} model is employed to extract visual features from all character images. Additionally, we compute cosine similarities between two characters based on their image features, and the maximum similarity value is used as the final similarity between the two characters. Finally, we apply the softmax function to convert the similarity values between a single character and all other characters into a probability distribution over replacement characters. To enhance character differentiation, we add a temperature parameter $\tau$ ($\tau=0.02$) into the softmax function, which promotes the character's likelihood of being replaced by a similar-looking character. 
% GPT
We first create multiple character images using various fonts and angles, encompassing both uppercase and lowercase formats for English letters. Next, we utilize a pre-trained ResNet50 model \cite{he2016resnet} to extract visual features from all character images. Subsequently, we compute cosine similarities between characters based on their image features, using the maximum similarity value as the final similarity. Finally, we use the softmax function to convert the similarity values into a probability distribution over replacement characters. To enhance character differentiation, we introduce a temperature parameter $\tau$ ($\tau=0.02$) into the softmax function, promoting the likelihood of a character being replaced by a visually similar one.
This matrix serves as a foundation for determining the likelihood of selecting specific replacement characters during the substitution process. By incorporating visual similarity information, SSLC simulates real-world annotation errors more accurately.

% orginal

\smallskip
\textbf{Online Corruption.} During training, a label undergoes online corruption twice using SSLC to generate two noisy labels. These labels and the original image form negative samples for training. During the corruption process, we randomly select 1 to 2 corruption operations and apply them in a specific order: deletion, substitution, transposition, and insertion. This order ensures that these operations do not affect each other. To account for the higher likelihood of substitution errors in real data and the challenge in identifying this noise, we assign a 30\% probability to select this corruption operation, while the probabilities for other operations are 20\% each. In \cref{fig:SELECT_results}, we display some corrupted labels after SSLC corruption. It can be seen that the lengths of some labels may change after SSLC corruption and substitution corruption tends to replace more similar characters.

% original [wait list]
The SSLC technique differentiates itself by capturing a wide array of errors, including insertion, deletion, substitution, and transposition. By incorporating a noise transition matrix, SSLC effectively simulates annotator confusion, prioritizing visually similar character substitutions. This mirrors the nuanced challenges of real-world scene text annotation and positions SELECT to address these with unprecedented accuracy.

% The SSLC technique distinguishes itself by capturing  a broad spectrum of error dynamics, encompassing insertion, deletion, substitution, and transposition. Through the integration of a noise transition matrix, SSLC effectively simulates annotator confusion, prioritizing visually similar character substitutions. This mirrors the nuanced challenges of real-world scene text annotation, positioning SELECT to address them with unparalleled accuracy.

% GPT
% The SSLC technique captures various error dynamics (insertion, deletion, substitution, and transposition) and incorporates a noise transition matrix to simulate annotator confusion, prioritizing visually similar character substitutions. This mirrors real-world scene text annotation challenges and positions SELECT to address them with unprecedented accuracy.

\begin{table*}[t]
    \centering
    \caption{Evaluation accuracy of three \textbf{STR} models on benchmarks using various methods to predict noisy labels and then remove same amount (24K) images before training.}
    \label{tab:SELECT_detect_real_results}
    \begin{tabular}{lllllllll}
    \toprule
     & & \multicolumn{7}{c}{\textbf{Dataset name and number of data}} \\
     \cmidrule(lr){3-9}
    
    \multirow{2}{*}{Model} &\multirow{2}{*}{Method} & IIIT & SVT & IC13 & IC15 & SP & CT & Total \\
    & & 3000 & 647 & 1015 & 2077 & 645 & 288 & 7672 \\
    \midrule
    
    \multirow{4}{*}{TRBA} & baseline & 95.90\% & 92.64\% & 95.25\% & 81.78\% & 85.77\% & \textbf{92.43\%} & 90.74\% \\ 
    ~ & Random & 95.73\% & 92.95\% & 95.05\% & \textbf{82.11\%} & 85.71\% & 91.39\% & 90.71\% \\
    ~ & PARSeq & 95.73\% & 92.05\% & 94.91\% & 81.26\% & 85.74\% & 90.76\% & 90.37\% \\ 
    ~ & CSL & 95.79\% & \textbf{93.20\%} & \textbf{95.27\%} & 81.51\% & 85.80\% & 91.67\% & 90.64\% \\ 
    ~ & SELECT & \textbf{96.25\%} & 92.74\% & 94.92\% & 82.04\% & \textbf{86.33\%} & 92.36\% & \textbf{90.95\%} \\ 
    \hline
    
    \multirow{4}{*}{VITSTR} & baseline & 94.37\% & 91.13\% & 94.23\% & 79.04\% & 83.35\% & 91.25\% & 88.89\% \\ 
    ~ & Random & 94.35\% & 91.04\% & 94.82\% & 78.68\% & 83.04\% & 90.62\% & 88.80\% \\
    ~ & PARSeq & 94.09\% & 90.94\% & 94.30\% & 78.76\% & 82.95\% & 90.62\% & 88.63\% \\ 
    ~ & CSL & 94.80\% & \textbf{92.33\%} & 95.05\% & 79.93\% & 83.88\% & 91.25\% & 89.55\% \\ 
    ~ & SELECT & \textbf{95.28\%} & \textbf{92.33\%} & \textbf{95.21\%} & \textbf{80.28\%} & \textbf{84.19\%} & \textbf{91.53\%} & \textbf{89.89\%} \\ 
    \hline
    
    \multirow{4}{*}{PARSeq} & baseline & 96.21\% & 94.59\% & \textbf{95.94\%} & 82.76\% & 87.54\% & 92.08\% & 91.51\% \\ 
    ~ & Random & 96.40\% & 94.47\% & 95.47\% & 82.50\% & 87.97\% & 92.29\% & 91.49\% \\
    ~ & PARSeq & 96.09\% & 94.03\% & 95.73\% & 82.25\% & 87.78\% & 92.01\% & 91.27\% \\ 
    ~ & CSL & 96.32\% & \textbf{94.71\%} & 95.82\% & 83.00\% & 87.88\% & 92.36\% & 91.66\% \\ 
    ~ & SELECT & \textbf{96.78\%} & \textbf{94.71\%} & 95.71\% & \textbf{83.14\%} & \textbf{88.40\%} & \textbf{93.54\%} & \textbf{91.95\%} \\ 
    \bottomrule
    \end{tabular}
  \end{table*}

\medskip
\section{Experiments}
\label{sec:experiments}

% This section first discusses the experimental setup and then present results on synthetic datasets are presented. Subsequently, we show the effectiveness of our method on the real STR datasets. Finally, we conduct ablative studies to explore the impact of various choices, including tokenizer selection, label corruption strategies, and auxiliary learning techniques.  

\subsection{Experimental Setup}

\smallskip
\textbf{Datasets.}
\label{sec:datasets}
To validate the effectiveness of our SELECT model, we first train it on SynthText (ST) \cite{gupta2016synthtext} and perform LED experiments on a manually corrupted synthetic dataset MJSynth (MJ) \cite{jaderberg2014mjsynth} to obtain direct quantitative results. Furthermore, we conduct experiments on real datasets, training SELECT on the ST and MJ datasets, and then detecting errors in the following 11 real datasets: SVT \cite{wang2011svt}, IIIT \cite{mishra2012iiit}, IC13 \cite{karatzas2013ic13}, IC15 \cite{karatzas2015ic15}, COCO-Text (COCO) \cite{veit1601cocotext}, RCTW \cite{shi2017rctw}, Uber \cite{zhang2017uber}, ArT \cite{chng2019art}, LSVT \cite{sun2019lsvt}, MLT19 \cite{nayef2019mlt19}, and ReCTS \cite{zhang2019rects}.
% Validating the accuracy of LED results on real datasets poses challenges due to the difficulty and cost of re-annotating the data. 
Moreover, noise often persists despite manual re-annotation. Inspired by Confident Learning \cite{northcutt2021confidentlearning}, we remove samples predicted as label errors by SELECT from real datasets and train the STR task in the resulting clean datasets. We aim to demonstrate how SELECT can be leveraged to attain comparable or improved results with reduced training data, indirectly showcasing the efficacy of the proposed approach. 
Specifically, we conduct these experiments on a combined dataset consisting of the 11 aforementioned real datasets, comprising 273K (after removing invalid data) labeled images.
For the evaluation benchmark of STR, we follow the previous work \cite{baek2021what} and use the following set of 6 datasets, totaling 7672 test images: IIIT \cite{mishra2012iiit}, SVT \cite{wang2011svt}, IC13-1015 \cite{karatzas2013ic13}, IC15-2077 \cite{karatzas2015ic15}, SVTP \cite{phan2013svtp}, and CUTE \cite{risnumawan2014cute}.

\medskip
\textbf{Training Protocol.}
All experiments are conducted using 8 NVIDIA Tesla A30 GPUs with PyTorch. We set a maximum label length of 25 and utilize a 36-character charset of lowercase alphanumeric characters.

% \subsubsection{SELECT} 
\smallskip
\textbf{\textit{1) SELECT:}} We train the SELECT model for ten epochs using a batch size of 512 and AdamW optimizer. % \cite{loshchilov2017adamw} with a weight decay of 0.05. 
The initial learning rate is 1e-5, decaying by 0.7 every epoch. A warm-up strategy is employed for 3000 steps with a learning rate of 1e-7. We use pre-trained BLIP \cite{li2022blip} weights for SELECT, excluding positional and patch embeddings due to altered image size. 
% We maintain a momentum-based model by computing moving average of parameters during training. 
We maintain a momentum-based model during training.
Image preprocessing involves resizing text images to $64\times256$ and applying RandAugment. %\cite{cubuk2020randaugment}. %for augmentation.
%
% \subsubsection{STR models} 

\smallskip
\textbf{\textit{2) STR models:}} For STR task, we train TRBA \cite{baek2021what}, ViTSTR \cite{atienza2021vitstr}, and PARSeq \cite{bautista2022parseq} using consistent hyperparameters: image size $32\times128$, learning rate 1e-3, batch size 64, and AdamW optimizer. % \cite{loshchilov2017adamw}. 
Training lasts 50 epochs, with SWA \cite{izmailov2018swa} applied during the final 15 epochs. All models are trained from scratch, except ViTSTR, which uses pre-trained weights from DeiT \cite{touvron2021deit} and applies RandAugment. % \cite{cubuk2020randaugment}. % for augmentation.

\smallskip
\textbf{Evaluation Protocol and Metrics.}
% \subsubsection{Metrics for label errors detection.}
For LED, we use precision, recall, and F1-score as metrics to evaluate the performance of label error detection in MJ test dataset.
% \subsubsection{Metrics for STR.} Word accuracy, which measures the correctness of predictions by considering all characters at all positions, is utilized as the primary metric for the STR evaluation benchmark. 
For STR, word accuracy is utilized as the primary metric, which measures the correctness of predictions by considering all characters at all positions.
A prediction is considered correct only if all characters match exactly. %To facilitate comparison, we calculate the total accuracy by combining the results from the six benchmark datasets, comprising 7,672 samples. 
Throughout all experiments, we conduct five trials with different initializations and report the average accuracy across all trials.

\subsection{Detecting Label Errors in Synthetic Datasets}

\begin{table}
    \centering
      \caption{Performance comparison of \textbf{Label Error Detection} methods on corrupted MJ test dataset.}
      \label{tab:corrupted_MJ}
      \centering
      \begin{tabular}{@{}llll@{}}
        \toprule
        Method & Precision & Recall & F1 \\
        \midrule
        PARSeq \cite{bautista2022parseq} & 89.24\% & \textbf{99.17\%} & 93.94\% \\
        CSL \cite{liu2021confidentsquencelearning} & 94.30\% & 97.38\% & 95.82\% \\
        SELECT & \textbf{98.33\%} & 98.58\% & \textbf{98.45\%} \\
      \bottomrule
      \end{tabular}
\end{table}

To validate our approach, we train SELECT on ST dataset and then detect label errors in the corrupted MJ test dataset, comprising 892K text images. We introduce 4 types of noise in equal proportions, corrupting 50\% of the MJ test set.

We employ the following two algorithms as baselines: (1) \textbf{PARSeq \cite{bautista2022parseq}}: This recent advancement is known as SOTA in STR. It is trained on the ST dataset and its predictions on the test set are compared against annotated texts to identify discrepancies as label errors for evaluation. 
% using a 7e-4 learning rate, a batch size of 384, and a duration of 20 epochs. 
% During evaluation, predictions on the test dataset are compared against annotated texts to identify discrepancies as label errors. 
(2) \textbf{Confident Sequence Learning (CSL) \cite{liu2021confidentsquencelearning}}: In this approach, we replace the original STR framework with the PARSeq model and then apply the CSL algorithm to detect label errors.
The performance metrics in \cref{tab:corrupted_MJ} provide an overview of each method's efficacy. Results show that PARSeq achieves a remarkably high recall rate of 99.17\%, slightly surpassing SELECT by 0.59\%, but with a lower precision at 89.24\%, 9.09\% less than that of SELECT, resulting in an F1 score 4.51\% lower than SELECT. CSL improves upon these results, achieving an F1 score of 95.82\%, demonstrating robust performance in accurately identifying label errors. Our proposed method, SELECT, surpasses all baseline algorithms, attaining the highest F1 score of 98.45\% and outperforming CSL across all metrics, demonstrating its superiority in detecting label errors in STR datasets. 
% These results unequivocally demonstrate the superior performance of SELECT in detecting label errors, outperforming baseline methods across all metrics.

\begin{table}
    \caption{Evaluation results for \textbf{Label Error Detection} using SELECT: 1.under various corruption strategies (CS) and 2.with or without auxiliary learning (AL) on the corrupted MJ test dataset.}
    \label{tab:label_corruption_results}
    \centering
    \begin{tabular}{l@{\hskip 0.3cm}l@{\hskip 0.3cm}l@{\hskip 0.3cm}l@{\hskip 0.3cm}l}
    \toprule
    CS & AL & Precision & Recall & F1 \\
    \midrule
    COBS & w/o & \textbf{99.79\%} & 92.63\% & 96.08\% \\
    SSLC & w/o & 96.55\% & 98.39\% & 97.46\% \\
    SSLC & w/ & 98.33\% & \textbf{98.58\%} & \textbf{98.45\%} \\
    \bottomrule
    \end{tabular}
\end{table}

\subsection{Detecting Label Errors in Real Datasets}

% As mentioned in \cref{sec:datasets}, evaluating the performance of label errors detection in real datasets is difficult. Inspired by Confident Learning \cite{northcutt2021confidentlearning}, we obtain a clean dataset by removing noisy label samples predicted by SELECT from the real datasets. We aim to explore whether training the STR models on the reduced clean datasets can achieve similar or higher recognition accuracy. To demonstrate the effectiveness of our method, we compare it with the aforementioned baselines, PARSeq \cite{bautista2022parseq} and CSL \cite{liu2021confidentsquencelearning}. We include an additional baseline of random sample removing to compare with other methods that remove the same number of samples. For fair comparison, we use these methods along with SELECT to each remove certain numbers of images from the dataset.

\begin{figure}
  \centering
  \begin{subfigure}{0.45\linewidth}
    \centering
    \includegraphics[width=1.0\linewidth]{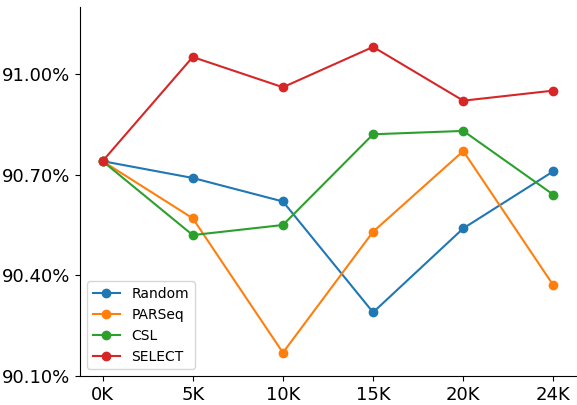}
    \caption{TRBA}
    \label{fig:trba}
  \end{subfigure}
  \begin{subfigure}{0.45\linewidth}
    \centering
    \includegraphics[width=1.0\linewidth]{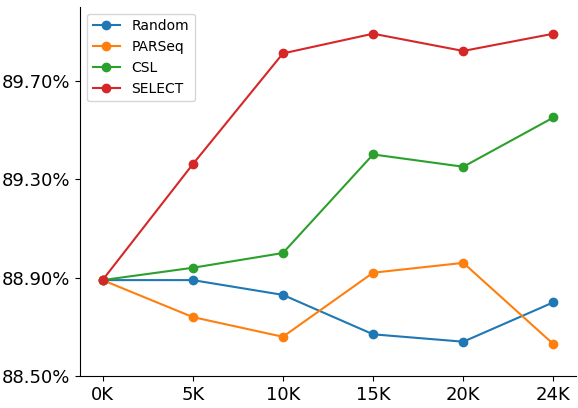}
    \caption{VITSTR}
    \label{fig:vitstr}
  \end{subfigure}
  \begin{subfigure}{0.45\linewidth}
    \centering
    \includegraphics[width=1.0\linewidth]{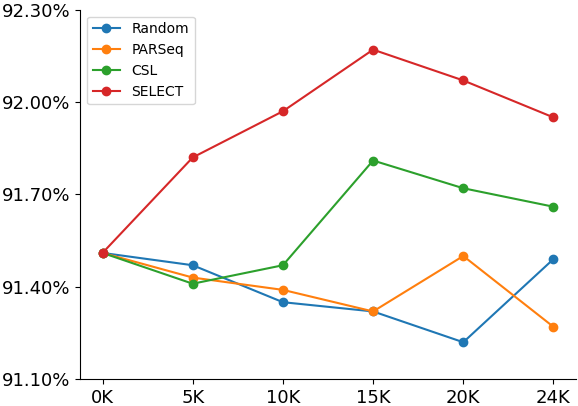}
    \caption{PARSeq}
    \label{fig:parseq}
  \end{subfigure}
  \caption{Evaluation accuracy of three \textbf{STR} models on benchmarks when using different methods to remove increasing numbers of noisy labels prior to training.}
    \Description{Evaluation accuracy of three \textbf{STR} models on benchmarks when using different methods to remove increasing numbers of noisy labels prior to training.}   
  \label{fig:SELECT_detect_real_results}
\end{figure}

To demonstrate the effectiveness of our method in real datasets, we follow the method \cite{northcutt2021confidentlearning} described in \cref{sec:datasets} and compare it with the aforementioned baselines, PARSeq \cite{bautista2022parseq} and CSL \cite{liu2021confidentsquencelearning}. We include an additional baseline of random sample removal to compare with other methods that remove the same number of samples. For a fair comparison, we use these methods along with SELECT to remove certain numbers of images from the dataset.
%
% \cref{fig:SELECT_detect_real_results} shows that SELECT achieves higher recognition accuracy compared to all baselines removal at all levels of sample removal using all three models, validating the effectiveness of SELECT. 
% While most of the STR accuracy for the three baselines decreased after removing different numbers of samples compared to the accuracy of baselines, the performance of SELECT after removing consistently increased: the highest accuracy of TRBA, VITSTR, and PARSeq increased by 0.35\%, 1.00\%, and 0.65\% respectively compared to the baselines. This demonstrates SELECT's efficiency not just in LED but also in maintaining or even enhancing model performance. This result indicates that training on clean datasets can improve the model's recognition accuracy even with a reduction of 8.7\% in noise samples. It also suggests that training STR tasks on existing noisy real datasets not only slows down model training convergence but also reduces recognition accuracy. \cref{tab:SELECT_detect_real_results} shows the details of the performance with a removal number of 24K (24K is the minimum of the predicted noise quantities by three methods in the prediction dataset). SELECT outperforms CSL by 0.31\%, 0.34\% and 0.29\% in accuracy for TRBA, VITSTR and PARSeq respectively.  
%
\cref{fig:SELECT_detect_real_results} shows that SELECT achieves higher recognition accuracy compared to all baseline removals across all levels of sample removal using the three models, validating its effectiveness. 
While most baseline STR accuracies decreased after sample removal, SELECT's performance consistently increased, with the highest accuracy of TRBA, VITSTR, and PARSeq improving by 0.35\%, 1.00\%, and 0.65\% respectively. This demonstrates SELECT's efficiency not only in LED but also in maintaining or even enhancing model performance. The results suggest that training on clean datasets can improve recognition accuracy, even with an 8.7\% reduction in noisy samples, and that training on noisy real datasets can slow model convergence and reduce accuracy for STR tasks. \cref{tab:SELECT_detect_real_results} details performance with a removal of 24K (the minimum predicted noise quantity by three methods in the prediction dataset), showing that SELECT outperforms CSL by 0.31\%, 0.34\%, and 0.29\% in accuracy for TRBA, VITSTR, and PARSeq respectively.
\cref{fig:SELECT_results} presents a visual representation of several noise samples detected by SELECT, showcasing its ability to identify four different types of noise: insertion, deletion, substitution, and transposition. It is also capable of detecting noise samples that contain multiple types of noise simultaneously, shown in \cref{fig:SELECT_results}(e).

\medskip
\subsection{Ablation Studies}
\label{sec:tokenizer_experiments}

% \smallskip

We conduct ablation studies to confirm the effectiveness of the choice of tokenizer and label corruption strategy in SELECT. For all experiments, we train the modified SELECT models on the ST dataset and detect label errors in the corrupted MJ test dataset.

\smallskip
\textbf{Tokenizer.} We study the effect of using a character-level tokenizer instead of WordPiece. We find that the model using WordPiece struggles to detect label errors, misidentifying most data as noise, with a low precision of 50.19\% and F1 of 66.81\%. This indicates that the WordPiece tokenizer, as used in the original multi-modal models, is unsuitable for character-level error detection in this task.

% \textbf{Label Corruption} In \cref{tab:label_corruption_results}, we compare two corruption strategies: Similarity-based Sequence Label Corruption (SSLC) versus corruption only by substitution (COBS) \cite{liu2021confidentsquencelearning}. It is observed that the model trained with the COBS accidentally performs well (thanks to the design of the architecture of SELECT), but there is still a gap compared to that with SSLC. The model trained with the COBS can detect insertion noise labels with high precision when the additional character(s) are inserted within the label. However, the model fails when the insertion occurs at the beginning or the end of the label, even when many characters are inserted in the label, as shown in \cref{fig:label_corruption_failed_cases} (a).This is because COBS treats internal insertions as substitutions, making them detectable, whereas it fails to locate spatial positions for characters inserted at the beginning or the end. Additionally, the COBS model poorly detects deletions, even when many characters are missing, as seen in \cref{fig:label_corruption_failed_cases} (b), and fails in cases of similar substitutions, shown in \cref{fig:label_corruption_failed_cases} (c). However, all of these examples can be detected by the SELECT model trained with SSLC. Consequently, the COBS model's recall is 5.76\% lower, and its F1 score is 1.38\% lower than the SSLC model's. These results indicate the effectiveness and importance of simulating the distribution of real data noise and employing similarity-based substitution during training.

\begin{figure}[h]
    \centering    
    \includegraphics[width=\linewidth]{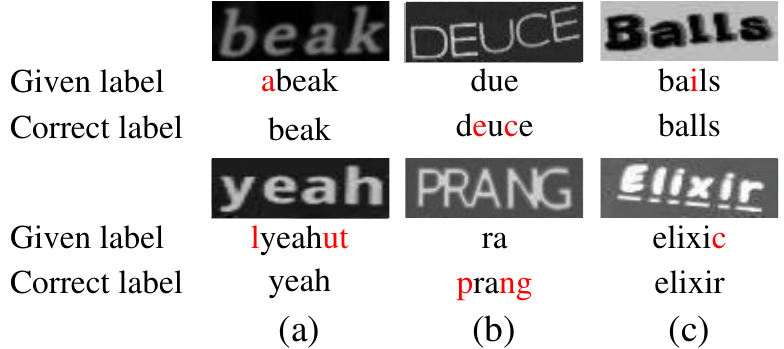}
    \caption{Failed cases of COBS corruption strategy. From left to right: (a) insertion, (b) deletion, (c) substitution. The incorrect or deleted characters in the labels are highlighted in red.}
    \Description{Failed cases of COBS corruption strategy. From left to right: (a) insertion, (b) deletion, (c) substitution. The incorrect or deleted characters in the labels are highlighted in red.}
    \label{fig:label_corruption_failed_cases}
\end{figure}

\smallskip
\textbf{Label Corruption.} In \cref{tab:label_corruption_results}, We compare two corruption strategies: SSLC and corruption only by substitution (COBS) \cite{liu2021confidentsquencelearning}. While the COBS-trained model accidentally performs well due to SELECT's architecture design, a performance gap persists compared to the SSLC-trained model. Specifically, the model trained with COBS exhibits high precision in detecting insertion noise labels when additional character(s) are inserted within the label. However, it fails when insertions occur at the label's beginning or end, even with multiple inserted characters, as depicted in \cref{fig:label_corruption_failed_cases} (a). This failure stems from COBS treating internal insertions as substitutions, rendering them detectable, while struggling to locate spatial positions for characters inserted at the beginning or the end. Additionally, the COBS model poorly detects deletions, even with multiple missing characters, as shown in \cref{fig:label_corruption_failed_cases} (b), and fails in cases of similar substitutions, shown in \cref{fig:label_corruption_failed_cases} (c). However, all of these examples can be detected by the SELECT model trained with SSLC. Consequently, the COBS model's recall is 5.76\% lower, and its F1 score is 1.38\% lower than the SSLC model's. These results indicate the effectiveness and importance of simulating the distribution of real data noise and employing similarity-based substitution during training.

\medskip
% \vsapce{1pt}
\textbf{Auxiliary Learning.} \cref{tab:label_corruption_results} also illustrates the influence of auxiliary learning (i.e., scene text recognition head) on SELECT's training results. The table reveals that the model with auxiliary learning exhibits a noteworthy enhancement across all four metrics compared to the model without it, showing a 0.99\% increase in the F1 score. The results indicate that the method not only effectively promotes the model's learning of text features within images but also helps the model become more attuned to character positions within the image, ultimately improving the accuracy of label error detection.

\vspace{-2pt}
\section{Conclusion}
\label{sec:conclusion}
% We present SELECT, the pioneering work in real-world scene text data label error detection (LED). Through image-text multi-modal training on synthetic datasets, a character-level tokenizer, and a Similarity-based Sequence Label Corruption technique, SELECT adeptly identifies label errors in real-world data. The experiment results showcase its ability to detect various types of noise, boosting accuracy for scene text recognition (STR) models. SELECT establishes a new benchmark for LED in STR, paving the way for multi-modal learning and error detection research.
We present SELECT, the pioneering work in real-world scene text data label error detection. Through image-text multi-modal training on synthetic datasets, a character-level tokenizer, and a Similarity-based Sequence Label Corruption technique, SELECT adeptly identifies label errors in real-world scene data, boosting accuracy for scene text recognition models. Results showcase its ability to detect various types of noise, enhancing model accuracy on real datasets. Ablation studies confirm the proposed components' effectiveness. SELECT establishes a new benchmark for label error detection in scene text recognition, paving the way for multi-modal learning and error detection research.

\bibliographystyle{ACM-Reference-Format}
\bibliography{select}
\end{document}